\documentclass{ifacconf}

\usepackage{graphicx}      % include this line if your document contains figures
\usepackage{natbib}        % required for bibliography
\usepackage[T1]{fontenc}
\usepackage{subcaption}

\usepackage{tikz}
\usetikzlibrary{shapes, arrows, positioning, calc, arrows.meta}

\usepackage{booktabs}
\usepackage{amssymb}
\usepackage{multirow}
\newcommand{\cmark}{\ensuremath{\checkmark}}
\newcommand{\xmark}{\ensuremath{\times}}

\usepackage{enumitem}
\usepackage{dblfloatfix}

\begin{document}
\begin{frontmatter}

\title{Seeing above the waves: A modular sensing framework for data acquisition at sea} 
%\thanksref{footnoteinfo}} 

\author[DTU]{Jonathan E. Schmidt} 
\author[DTU]{Julius Wirbel}
\author[DTU]{P. Nicholas Hansen}
\author[DTU]{Morgan Louédec}
\author[DTU]{Christian L. H. Westerdahl}
\author[DTU]{Dimitrios Dagdilelis}
\author[DTU]{Roberto Galeazzi}

\address[DTU]{Technical University of Denmark, Kongens Lyngby, Denmark (e-mail: \{jonei, roga\}@dtu.dk).}

\begin{abstract}
Advancing autonomy for surface vessels requires systematic evaluation of their sensing and perception subsystems. Yet, maritime environments impose unique challenges: sensor installation is constrained by vessel layout, environmental conditions such as fog or sea clutter are difficult to reproduce, and long-duration missions complicate data collection. This work addresses the question: How can we design a modular and reproducible sensor platform for maritime autonomy?

We present a comprehensive design blueprint that incorporates diverse modalities - RADAR, LiDAR, IMU, GNSS, AIS, RGB and LWIR cameras, and weather sensors - to enhance environmental awareness and vessel proprioception. Supported by a dedicated ROS2-based software framework for data management, our modular platform enables long-term data collection, hardware-in-the-loop testing, and integration with existing sensors and algorithms.
By unifying hardware design and data capture methodology, the platform enhances reproducibility and comparability across vessels and research projects. The proposed framework bridges engineering implementation and research methodology, providing the foundation for standardized, verifiable datasets essential to advancing situational awareness and autonomous maritime navigation.
\end{abstract}


\end{frontmatter}
%===============================================================================

\section{Introduction}
Research in maritime autonomy is often inhibited by limited or fragmented datasets, as data is frequently not publicly available~\citep{mirza_robustness_2021, wirbel2025data}. This lack of accessible data restricts progress in the field and underscores the need for systematic data collection efforts. To address this, we propose a standardized sensor setup that facilitates the creation of diverse datasets.

A common framework for data collection and storage accelerates deployment and simplifies data sharing among researchers. \citet{thombre2022} outlined operational requirements for maritime autonomy and evaluated sensor modalities accordingly. We extend these insights by presenting a practical configuration for sensor integration and data collection.

Standardization not only facilitates cross-comparison between vessels and operational areas but also enhances collaboration among research groups through a unified data format. Additionally, it reduces setup and development time for new deployments.
In particular, structured multimodal recordings enable reproducible evaluation of perception and state-estimation pipelines, while supporting hardware-in-the-loop validation of decision-making algorithms.

In this paper, we aim to bridge the gap between engineering practice - designing robust maritime sensor platforms - and research methodology - producing standardized, verifiable datasets. We argue that such infrastructure is essential for advancing situational awareness and autonomy in surface vessels.

To address this multitude of requirements, we propose a modular sensor platform blueprint that emphasizes flexibility, reproducibility, and long-term operability, designed to support multimodal sensing, and scalable data collection. We aim to provide a blueprint that can be easily adopted and modified by the community to develop their own sensor platforms. We further illustrate the adaptability of the platform through examples from experimental campaigns made possible by the architecture and related functionality.

\section{Related work}
Work towards autonomous surface vessels has gained more traction within recent years, driven by the advance in technology, regulatory interest, and the pursuit of increased safety onboard manned vessels. Therefore, various research projects have set sail to explore and refine autonomy in maritime contexts - from decision support and situational awareness to fully unmanned navigation. These efforts have highlighted the critical role of sensors in reaching higher levels of autonomy.

\cite{blanke2024greenhopper} presented the development of the \emph{GreenHopper} test bed, introducing a holistic system architecture of an autonomous vessel including modules for situational awareness and navigation using various different sensors for exteroception and proprioception. The findings from this work highlight the importance of situational awareness from diverse sources, as well as the requirement for diverse datasets when training and deploying machine learning-based algorithms in a highly dynamic environment. Similarly, \cite{eide2025} proposed \emph{milliAmpere2}, a ferry designed to perform fully autonomous passages in urban waterways. While this project did not directly focus on data collection, the authors followed a similar approach to data collection and autonomy.

\cite{chung2023pohang} developed a data collection system including cameras, LiDAR, and GNSS, mounted on a cruise boat. To ensure time alignment across all sensor streams, the system employed precise time synchronization protocols. Additionally, sensor calibration metadata is provided alongside the dataset.

Following a slightly different approach, \cite{defilippo-robowhaler} developed the \emph{RoboWhaler}, a modified 7m Boston Whaler serving as a versatile marine autonomy laboratory to support algorithm development and on-water dataset collection for autonomous surface vessels. This platform can operate in both manned and unmanned modes. The vessel has been used in continuous multi-modal data gathering campaigns from 2019 through 2021, resulting in several terabytes of curated sensor data that are publicly available, and it remains in active use with yearly sensor upgrades – demonstrating its upgradability and value as a long-term test bed for marine autonomy research.

\cite{labust_sensorplatform} presented a modular sensor rack designed for rapid deployment on small, manned surface vessels, aiming to address the lack of high-quality multi-sensor maritime data needed for the development of reliable autonomous navigation systems. It was demonstrated in a two-day data collection campaign along the Croatian coast, capturing synchronized multi-modal data for model training. Built from aluminium modular profiles with a self-contained power supply, the system emphasizes easy installation on different vessel types.

Our work builds on the lessons learned from the approaches introduced above by introducing a system that is modular, easy to deploy on different vessel types and for different mission profiles while keeping collection centralised and synchronised for easy post-processing.

Table~\ref{tab:rw} provides a comparison of the reviewed platforms.

\renewcommand{\arraystretch}{0.9}
\begin{table*}[btp]
\centering
\caption{Comparison of related maritime sensing platforms.
         \cmark~= yes; \xmark~= no; {--}~= not reported. Cont. SW = containerised software}
\label{tab:rw}
\footnotesize
\setlength{\tabcolsep}{4pt}
\resizebox{\textwidth}{!}{%
\begin{tabular}{l ccccccc cc lc}
\toprule
\multirow{2}{*}{Platform}
  & \multicolumn{7}{c}{Sensor Modalities}
  & \multirow{2}{*}{\shortstack{Platform-\\Agnostic}}
  & \multirow{2}{*}{\shortstack{Cont.\\SW}}
  & \multirow{2}{*}{\shortstack{Data\\Format}}
  & \multirow{2}{*}{\shortstack{Public\\Dataset}} \\
\cmidrule(lr){2-8}
& RADAR & LiDAR & RGB & IR & IMU & GNSS & Additional & & & \\
\midrule
GreenHopper~\citep{blanke2024greenhopper}  & \cmark & \cmark & \cmark & \cmark & \cmark & \cmark & \cmark & \xmark & --     & --      & --      \\
milliAmpere2~\citep{eide2025}               & \cmark & \cmark & \cmark & \cmark & \cmark & \cmark & \cmark & \xmark & --     & --      & --      \\
Pohang~\citep{chung2023pohang}              & \cmark & \cmark & \cmark & \cmark & \cmark & \cmark & -- & \xmark & --     & Individual & Yes     \\
RoboWhaler~\citep{defilippo-robowhaler}     & \cmark & \cmark & \cmark & \cmark & \cmark & \cmark & -- & \xmark & --     & ROS bag & Yes     \\
Labust et al.~\citep{labust_sensorplatform} & --     & \cmark & \cmark & --     & \cmark & \cmark & \cmark & \cmark & --     & ROS bag   & Planned \\
\midrule
This work                                   & \cmark & \cmark & \cmark & \cmark & \cmark & \cmark & \cmark & \cmark & \cmark & MCAP   & Planned \\
\bottomrule
\end{tabular}
}
\end{table*}

% Styles
\tikzstyle{camera} = [rectangle, draw, rounded corners, text centered, minimum height=2em, minimum width=6em, fill=blue!20]
\tikzstyle{rangesens} = [rectangle, draw, text centered, minimum height=2em, minimum width=6em, fill=green!20]
\tikzstyle{radiosens} = [rectangle, draw, text centered, minimum height=2em, minimum width=6em, fill=orange!30]
\tikzstyle{aidingsens} = [rectangle, draw, text centered, minimum height=2em, minimum width=6em, fill=yellow!30]
\tikzstyle{extsens} = [rectangle, draw, text centered, minimum height=2em, minimum width=6em, fill=purple!30]
\tikzstyle{compute} = [rectangle, draw, text centered, minimum height=2em, minimum width=6em, fill=gray!20]
\tikzstyle{network} = [rectangle, draw, text centered, minimum height=2em, minimum width=6em, fill=white!20]

\tikzstyle{line} = [draw, thick]
\tikzstyle{usbline} = [draw, thick, color=red]  % choose your color (e.g. red, blue, cyan)
\tikzstyle{dot} = [circle, fill=black, minimum size=3pt, inner sep=0pt]

\begin{figure}[tbp]
\centering
\resizebox{\columnwidth}{!}{%
\begin{tikzpicture}[node distance=0.75cm]

% Main compute nodes
\node[compute] (jetson) {Jetson Xavier};

% RGB camera cluster (horizontal)
\node[camera, above=of jetson] (cameran) {RGB Camera n};
\node[dot, left=0.1cm of cameran] (dot1) {};
\node[dot, left=0.1cm of dot1] (dot2) {};
\node[dot, left=0.1cm of dot2] (dot3) {};
\node[camera, left=0.1cm of dot3] (camera1) {RGB Camera 1};

% IMU and AIS
\node[aidingsens, right=of cameran] (imu) {IMU};
\node[extsens, right=of imu] (ais) {AIS};

% Raspberry Pis and LWIR cameras (vertical)
\node[compute, right=of jetson] (rpi1) {Raspberry Pi};
\node[compute, below=of rpi1] (rpi2) {Raspberry Pi};

\node[camera, right=of rpi1, xshift=-0.2cm] (lwir1) {LWIR Camera 1};
\node[dot, below=0.1cm of lwir1] (vdot1) {};
\node[dot, below=0.1cm of vdot1] (vdot2) {};
\node[dot, below=0.1cm of vdot2] (vdot3) {};
\node[camera, right=of rpi2, xshift=-0.2cm] (lwirn) {LWIR Camera n};

% Network and environmental sensors
\node[network, below=of jetson] (poe) {PoE Switch};
\node[rangesens, left=of poe] (radar) {Radar};
\node[radiosens, above=of radar] (compass) {Satellite Compass};
\node[aidingsens, below=of poe] (weather) {Weather Station};
\node[rangesens, below=of radar] (lidar) {Lidar};
\node[rangesens, below=of rpi2] (sonar) {Sonar};

% --- Connections ---

% --- USB Connections ---
%\path[usbline] (camera1.south) to[out=-45,in=90]  (jetson.north);
%\path[usbline] (cameran.south) to[out=-90,in=0] (jetson.north);
%\path[usbline] (imu.south) to[out=-135,in=45] (jetson.north);
%\path[usbline] (ais.south) to[out=-120,in=20](jetson.north);
\path[usbline] (camera1.south) -- ++(0., -0.375) -- ++(3.4, 0) -- (jetson.north);
\path[usbline] (cameran.south) -- (jetson.north);
\path[usbline] (imu.south) -- ++(0., -0.375) -- ++(-3.14, 0) -- (jetson.north);
\path[usbline] (ais.south) -- ++(0., -0.375) -- ++(-6, 0) --(jetson.north);
\path[usbline] (lwir1) -- (rpi1);
\path[usbline] (lwirn) -- (rpi2);

% Network and other sensors
\path[line] (weather) -- (poe);
\path[line] (rpi1.south)-- ++(0., -0.375) -- ++(-3.06, 0) -- (poe.north);
\path[line] (poe) -- (rpi2);
\path[line] (jetson) -- (poe);
\path[line] (poe) -- (radar);
\path[line] (compass.south) -- ++(0., -0.375) -- ++(2.86, 0) -- (poe.north);

\path[line] (lidar.north) -- ++(0., 0.375) -- ++(2.86, 0) -- (poe.south);
\path[line] (sonar.north)-- ++(0., 0.375) -- ++(-3.06, 0) -- (poe.south);

\end{tikzpicture}
}
\vspace{-1.5em}
\caption{Example network diagram of the proposed platform. Box colours denote hardware type—computers (grey); imaging (blue); ranging (green); radio frequency (orange); aiding (yellow); external (pink); supporting (white)—with Ethernet (black) and USB (red) links; dots indicate multiple applicable sensors.}\label{fig:network_diagram}
\end{figure}
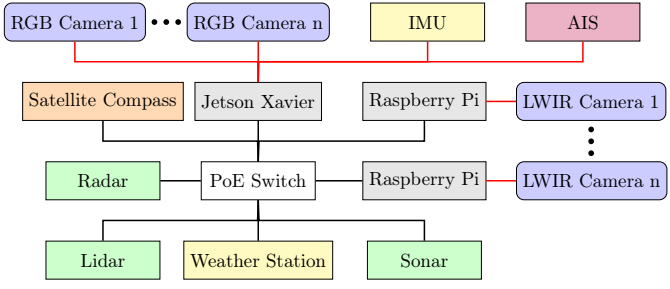

\section{A blueprint for a sensor platform}

Developing autonomous vessels requires a robust and adaptable sensor infrastructure capable of supporting diverse research goals, vessel types, and environmental conditions. Such a platform should offer the flexibility of accommodating various mission plans for a given research project while being generalisable enough that they can accommodate potential future campaign plans.

\subsection{Soft Requirements for the Platform}
Designing a platform that remains flexible across different vessel types and research missions requires a set of high-level, or \textit{soft}, requirements. These serve as guiding principles rather than strict specifications, favouring adaptability and reproducibility across implementations.

\begin{itemize}
    \item \textbf{Modularity:} Modularity is central to the blueprint, enabling rapid adaptation to different campaigns by modifying or extending sensor configurations without redesigning the overall system.
    \item \textbf{Hardware and Software Isolation:} Each functional element should be physically and logically isolated, using hardware separation to prevent cascading faults and software containerization to contain failures and improve reliability and maintainability.
    \item \textbf{Network Setup:} Networking underpins the distributed platform, requiring reliable yet simple communication between sensors, control nodes, and capture units, with static IPs for core nodes and dynamic connections for user devices to minimize routing overhead.
    \item \textbf{Platform Calibration:} Modular designs complicate calibration due to varying sensor configurations, requiring simple, repeatable routines linked to the data, which can be ensured by storing raw data and applying known calibrations post-collection.
    \item \textbf{Reproducibility and Longevity:} To promote long-term use, reproducibility, and collaboration, hardware and software interfaces should follow open, well-documented standards.
\end{itemize}

\subsection{Hardware Stack: A Modular, Multimodal System}
The hardware stack provides the physical foundation for the modular platform and is organized around three core elements: the \textit{Sensor Modalities}, the \textit{Sensor Capture Nodes} and the dedicated \textit{Data Collection Node}. This architecture enables multimodal data capture across heterogeneous sensor types, while maintaining scalability and resilience against single sensor failures. Figure~\ref{fig:network_diagram} shows an example network diagram of a modular sensor platform based on this blueprint.

\emph{Sensor Modalities -} To enable diverse research and development within the maritime domain, the platform supports multiple sensing modalities simultaneously. Drawing on prior work~\citep{thombre2022, dagdilelis2022, blanke2024greenhopper, eide2025, Louedec_2025, Schmidt_2025online, schmidt2025gabor}, the following categories were considered for maritime campaigns:

\begin{description}[leftmargin=5pt,labelindent=0pt]
    \item[Imaging Sensors] comprise visible and infrared cameras, enabling operation under varying conditions such as darkness or poor weather, with optional dedicated camera sets for vessel areas of specific research interest.
    
    \item[Ranging Sensors] provide relative range and speed information, using X-band RADAR for medium- to long-range navigation and validation, and LIDAR or K-/W-band RADAR for high-resolution short-range tasks such as tracking or docking.
    
    \item[Radio Frequency Sensors] include Global Navigation Satellite Systems (GNSS) receivers for position, speed, and course estimation, supplemented by satellite communication links for monitoring or alternative localization~\citep{thombre2022, grayver2024}.
    
    \item[Aiding Sensors] incorporate IMUs, gyroscopic compasses, and speed or depth sensors for motion estimation and inertial navigation, as well as weather sensors supplying environmental context affecting sensor performance and navigation.
      
    \item[External Information Systems] integrate navigational data from Electronic Chart and Information System (ECDIS) and vessel information from Automatic Identification System (AIS), alongside external weather and current reports for validation of onboard measurements.
\end{description}

%\begin{figure}[tbp]
%\begin{center}
%\includegraphics[scale = 0.3]{figures/docker-structure.png}    % The printed column width is 8.4 cm.
%\caption{Visualization of the Docker Container Architecture} 
%\label{fig:docker}
%\end{center}
%\end{figure}

\emph{Sensor Capture Nodes -} Capture nodes are distributed processing units, e.g. Raspberry PIs, co-located with the sensors, which manage local data capture and preliminary data handling. This reduces the load on the main network and the data collection node. Physical and logical isolation of these nodes provides fault containment: if one node fails, the others remain operational. Utilizing dedicated capture nodes for modalities like Imaging Sensors allows the placement of these sensors at up to 100 metres from the centrally located switch using Ethernet and Power Over Ethernet (PoE)+, enhancing coverage of relevant surrounding areas of the vessel.

\emph{Data Collection Node -} The dedicated data collection node serves as the central logging and system coordination unit of the platform. It handles the control commands to the capture nodes, time synchronization, network routing, and centralized data storage. Relying on a single, centralized node to perform these tasks simplifies the network design, but introduces a single point of failure. If a mission requires full redundancy, a second node running the same software stack can be added to serve as a live backup.

\emph{Sensor Hardware and Interfaces -} All components in this blueprint are connected using Ethernet due to its wide compatibility and robustness. PoE was chosen for simplicity and cost efficiency, allowing simultaneous power and data transmission over long distances. This allows for easy interfacing between the different, available sensor hardware and the main platform.

\subsection{Software Stack: Distributed and Containerized}
\label{sec:software}
The software architecture also implements the modularity and isolation principles introduced earlier. It builds on the Robot Operating System 2 (ROS2, ~\cite{Macenski_2022}) middleware for distributed communication, and Docker for containerized deployment.

    \emph{Distributed Communication Middleware -}
    ROS2 provides a flexible, message-based communication layer across heterogeneous hardware, which enables standardized integration of sensors with embedded controllers, microcontrollers, and single-board computers. Its mature ecosystem of open-source packages reduces development effort and encourages reuse, while allowing control commands, sensor data and local inference outputs to be exchanged on the same network using custom message types where needed.
    ROS2 also provides a reliable way of capturing raw data and associated transforms for sensor location and calibration through topic-based subscriptions, enabling distributed and redundant capture across the network.
    The platform uses CycloneDDS as the ROS2 middleware, with a fixed \texttt{ROS\_DOMAIN\_ID} to isolate traffic. Selected sensor topics use a \texttt{KEEP\_LAST} (depth~1) history policy with \texttt{TRANSIENT\_LOCAL} durability, ensuring that late-joining subscribers - such as the centralised recorder - always receive the most recent sample without unbounded queues.
    
\begin{figure}[tbp]
\centering
\vspace{-0.5em}
\resizebox{0.8\columnwidth}{!}{%
\begin{tikzpicture}[font=\sffamily, baseline]

% Layers
\node[draw, fill=blue!30, text=black, minimum width=6cm, minimum height=0.75cm] (hardware) {Physical Hardware};
\node[draw, fill=green!40, text=black, minimum width=6cm, minimum height=0.75cm, above=0cm of hardware] (os) {Operating System};
\node[draw, fill=magenta!40, text=black, minimum width=6cm, minimum height=0.75cm, above=0cm of os] (docker) {Docker Engine};

% Docker containers
\node[draw, fill=yellow!30, minimum width=2.5cm, minimum height=0.75cm, above=of docker.west, anchor=west, xshift=-0.0cm, yshift=-0.25cm] (container1) {Docker Container};
\node[draw, fill=yellow!30, minimum width=2.5cm, minimum height=0.75cm, above=of docker.east, anchor=east, xshift=0.0cm, yshift=-0.25cm] (container2) {Docker Container};

% ROS2 nodes
\node[draw, fill=blue!20, minimum width=2.8cm, minimum height=0.75cm, above=0cm of container1] (ros1) {ROS2 Node};
\node[draw, fill=blue!20, minimum width=2.8cm, minimum height=0.75cm, above=0cm of container2] (ros2) {ROS2 Node};

% % Sensors cloud
% \node[cloud, draw, cloud puffs=10, cloud puff arc=120, aspect=2, minimum width=3cm, minimum height=0.75cm, below=of hardware, yshift=0.75cm] (sensors) {Sensors};

% % Arrow from sensors to hardware
% \draw[->, thick] (sensors) -- (hardware);

% Isolation Level arrow
%\draw[<-, very thick] 
%  ([xshift=-1.5cm,yshift=0cm]ros1.north west) -- ([xshift=-1.5cm,yshift=-0.5cm]hardware.south west)
%  node[midway,left,align=center,draw=black,fill=white,inner sep=2pt] {Isolation Level};
\draw[
  -{Stealth[width=30pt,length=16pt]}, % big arrowhead
  line width=2pt,
  draw=black,
  fill=white,
  double=white, double distance=10pt, % white interior, black edge
  shorten >=1pt, shorten <=15pt
] ([xshift=-1.5cm,yshift=-0.5cm]hardware.south west)
  -- ([xshift=-1.5cm,yshift=0cm]ros1.north west)
  node[midway, rotate=90, text=black, align=center] {Isolation Level};

\end{tikzpicture}
}
\vspace{-2em}
\caption{Layered architecture of isolated ROS2 nodes in Docker containers, illustrating the isolation level between individual sensors in software.} 
\label{fig:docker}
\end{figure}
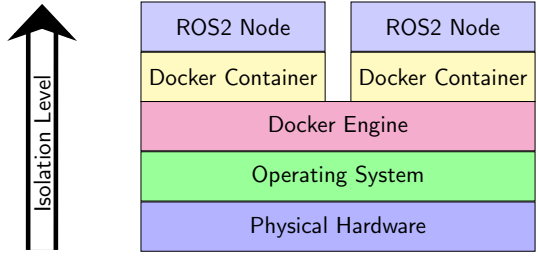

    \emph{Docker Isolation -}
    Each sensor runs on a dedicated node encapsulated in a minimal Docker container with only the required ROS2 components and modality-specific libraries, ensuring clean dependency management, simple deployment, and fault isolation. Hardware supporting the runtime environment can be used for deployment, allowing flexible integration of sensors and nodes. Figure \ref{fig:docker} illustrates the typical container composition.
    Containers share the host network namespace, enabling transparent DDS discovery without additional configuration, while host inter-process communication and shared memory reduce serialization overhead for high-throughput modalities such as imaging. Automatic container restarts provide lightweight fault recovery, and modality-specific services are orchestrated via Docker Compose, allowing sensors to be easily enabled or disabled per campaign.
    This containerized approach also facilitates sharing sensor capture and inference implementations across research groups, supporting collaboration and standardization.
    
    \emph{Internal Network Design -}
    Reliable networking forms the backbone of the software stack. A central control node manages IP routing and time synchronization across all nodes using an Gigabit PoE switch. Static IPs are assigned to sensor and capture nodes, while external monitoring or control devices receive dynamic addresses. A firewall isolates platform traffic from the vessel’s network and any external connectivity.
    To handle high-throughput sensors such as LiDAR or imaging cameras, the system supports two complementary strategies:
    
\begin{description}[leftmargin=5pt,labelindent=-5pt]
\item \textbf{Separate Networks:} Grouping high-frequency modalities into dedicated sub-networks to prevent data congestion.
\item \textbf{Local Capture:} Recording data directly at capture nodes to reduce network load and improve robustness to communication losses.
\end{description}

\subsection{Collection Methodology, Data Format, and Calibration}
The blueprint’s collection methodology aims to ensure consistent, high-quality data acquisition independent of vessel integration. The mission plan defines what data to collect, at what frequency, and in which format. The system’s flexibility allows dynamic configuration of these parameters to meet different campaign objectives.

\emph{Sampling and Synchronization -}
    Sensors operate at different frequencies and resolutions, requiring configurable sampling without data loss.  To ensure time consistency, all data are synchronized to a single \emph{platform time} derived from GNSS or a GNSS-aligned hardware real-time clock. The containerized architecture supports both continuous sampling at native sensor rates and pull-based acquisition, where capture commands trigger synchronized measurements across selected nodes.
    
\emph{Calibration and Sensor Layout -}
    Sensor calibration metadata, including intrinsic and extrinsic parameters,  are stored with the data to ensure accurate sensor layout for later fusion, mapping, and tracking across configurations and campaigns.
    
\emph{Data Capture Format and Storage -}
    The platform uses the ROS-native \texttt{MCAP} format to record all ROS2 messages in compliance with FAIR data principles, with precise header timestamps ensuring temporal traceability.  MCAP supports fast lossless compression (zstd) and visualization using ROS-native or third-party tools~\citep{chung2023pohang, choi2025polaris, jang2025moana}. Recordings are automatically split into fixed-duration segments  and stored on a dedicated external drive, while each sensor module is packaged with a Dockerfile and a Compose file.

\begin{figure*}[t]
    \centering
    \begin{subfigure}[b]{0.32\textwidth}
        \centering
        \includegraphics[width=\textwidth]{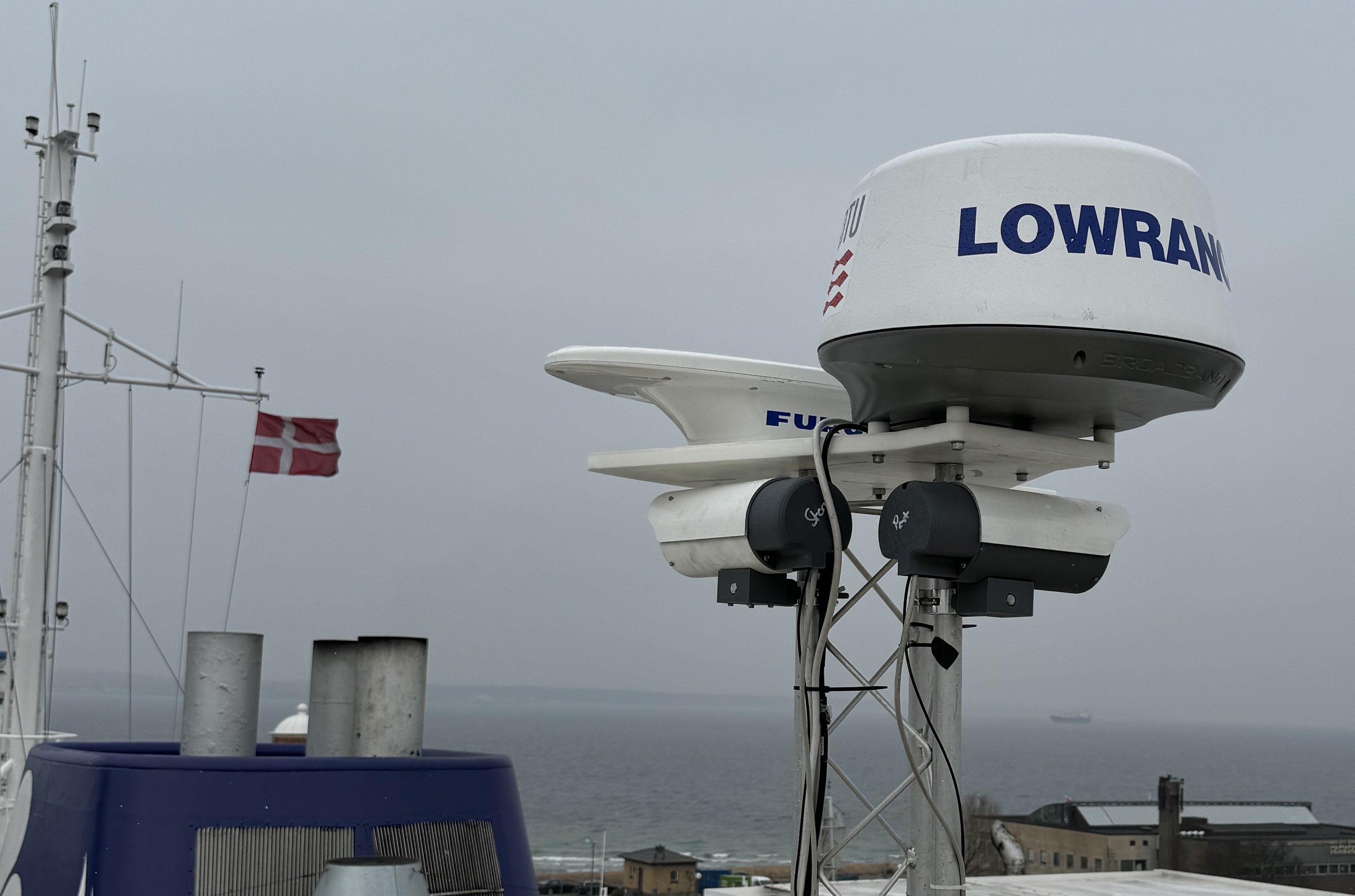}
        \caption{RADAR, satellite compass, and two camera houses with an RGB camera, an LWIR camera and a Raspberry Pi onboard \emph{Aurora}.}
        \label{fig:sub1}
    \end{subfigure}
    \hfill
    \begin{subfigure}[b]{0.32\textwidth}
        \centering
        \includegraphics[width=\textwidth]{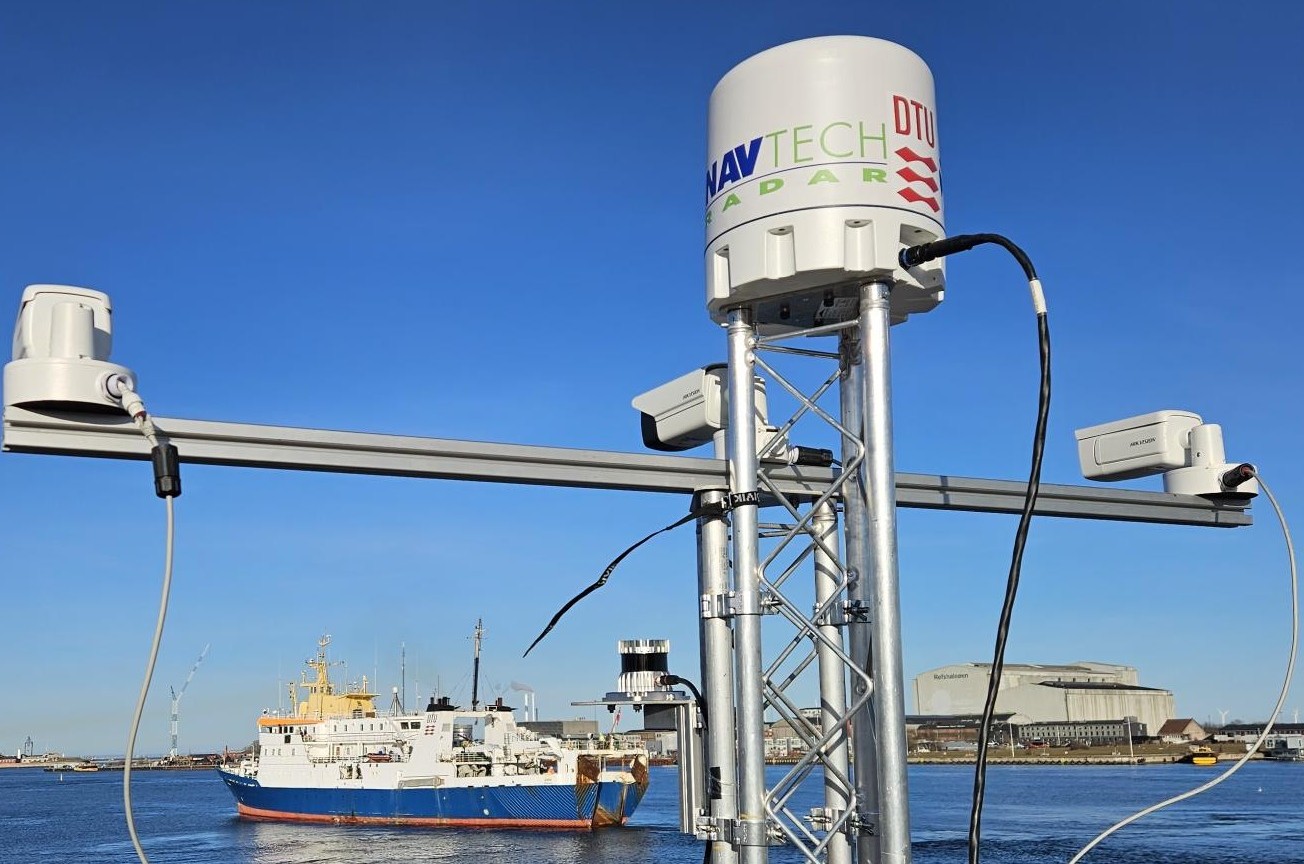}
        \caption{Millimeter wave RADAR, LiDAR, one LWIR camera, and two RGB cameras, located on land at the harbour bus.}
        \label{fig:sub2}
    \end{subfigure}
    \hfill
    \begin{subfigure}[b]{0.32\textwidth}
        \centering
        \includegraphics[width=\textwidth]{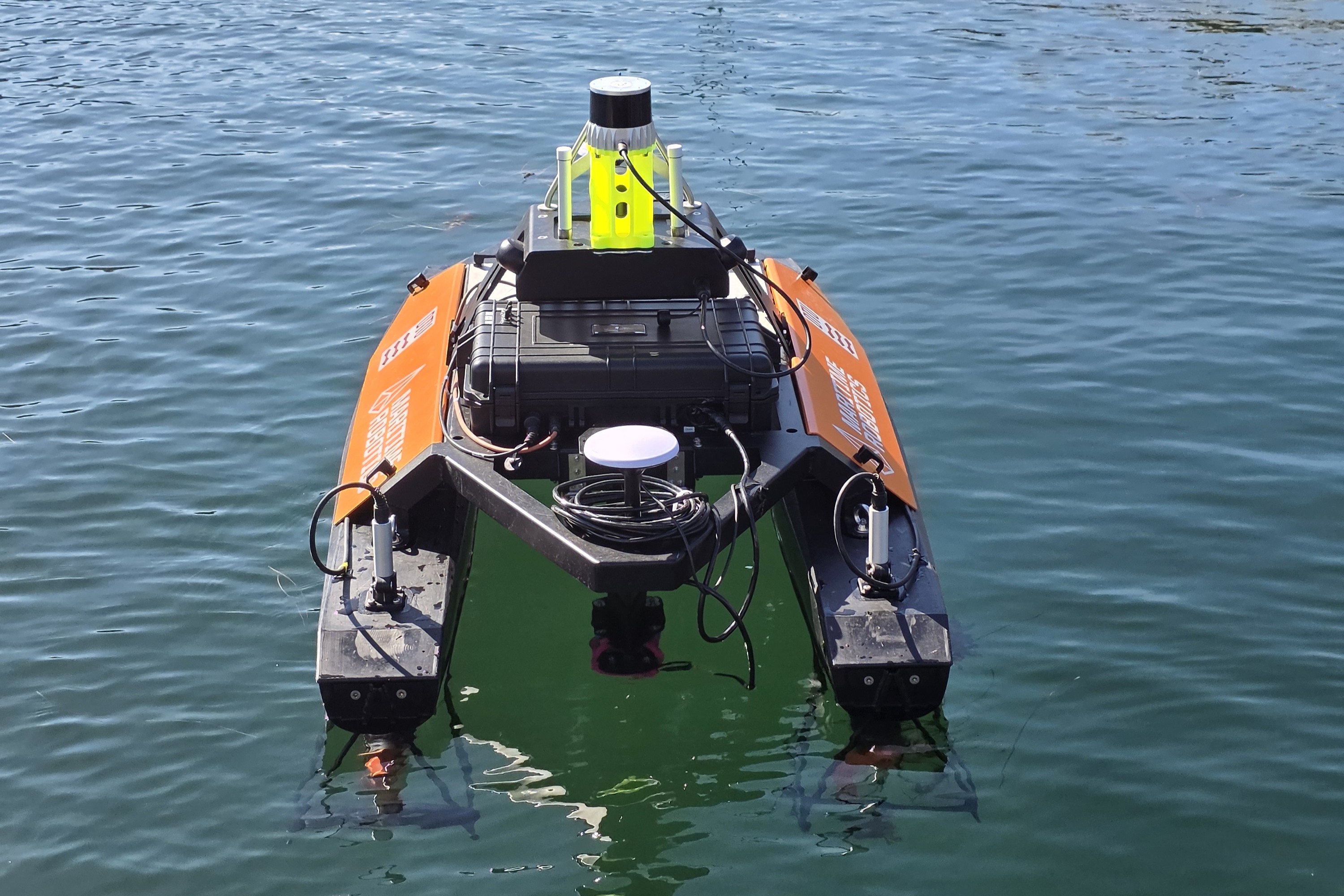}
        \caption{LiDAR and GNSS antenna mounted on top of a USV and a sonar mounted underneath.} %Not pictured but equipped: Two SONARs, and a 9-axis IMU.}
        \label{fig:sub3}
    \end{subfigure}
    \caption{Examples of (parts of) sensor platforms used in experimental campaigns. In each campaign all components are connected via Ethernet to a central collection node.}
    \label{fig:campaigns}
\end{figure*}

\section{Campaigns based on the blueprint}
Sensor platforms based on the proposed blueprint have been deployed across multiple data collection campaigns, providing valuable insights into the practical requirements for modularity and adaptability in maritime research. These campaigns have demonstrated the versatility of different sensor modalities and configurations, highlighting how sensor type and position can significantly influence the achievable situational awareness.

The experience gained through these deployments has informed refinements to the platforms, ensuring they remain suitable for a wide range of research objectives - such as autonomous navigation and environmental monitoring. Table~\ref{tab:campaigns} lists the key characteristics of each campaign.

\begin{table*}[!b]
\centering
\caption{Summary of data collection campaigns based on the proposed blueprint.
         Duration and data volume are reported per deployment.
         WS~= weather station.}
\label{tab:campaigns}
\footnotesize
\setlength{\tabcolsep}{4pt}
\begin{tabular}{p{3.5cm} c c c p{7.5cm}}
\toprule
Campaign & Deployments & Duration & Data Volume & Sensor Modalities \\
\midrule
Coastal navigation
  & 2 & 12--24 days & 4--5 TB
  & RADAR, RGB (4), LWIR (2--4), GNSS, IMU, WS (2) \\[3pt]
Urban transportation
  & 4 & 2--5 h & 100-500GB
  & Vessel: RADAR, LiDAR, RGB (4), LWIR (4), GNSS, IMU\newline
    Land: mmW-RADAR, LiDAR, RGB (2), LWIR \\[3pt]
Infrastructure mapping
  & 5+ & 3-7h & 100-500GB
  & LiDAR, GNSS, IMU, SONAR (1--2) \\
\bottomrule
\end{tabular}
\vspace{-1em}
\end{table*}

\subsection{Navigation in coastal waters}
A sensor platform was installed onboard the ferry \emph{Aurora}, which daily crosses the Øresund strait between Denmark and Sweden multiple times. Part of the platform is shown in Fig.~\ref{fig:campaigns}(a).

\emph{Mission Goals:} (i) Assess how changes in weather conditions influence the performance of perception sensors and downstream tasks; (ii) Validate the feasibility of long-duration sensor data acquisition using multiple modalities in a real-world maritime environment where weather, illumination and various maritime traffic can be captured.

\emph{Results:} Weather was identified as a key factor affecting sensor performance~\citep{Schmidt_2025online, schmidt2025gabor}, particularly for RGB cameras under different phenomena. The initial campaign highlighted the need for synchronized, structured data handling, motivating the adoption of ROS2. A second campaign with additional cameras and improved cabling confirmed the robustness and reliability of the refined hardware and software stack during extended maritime operations.

\subsection{Urban waterborne transportation}
The sensor platform, used onboard the ferry \emph{Aurora}, was also deployed on harbour buses operating in Copenhagen, accompanied by a sister platform installed at a landing quay. Due to the close proximity to shore, an Ouster LiDAR was added to the onboard platform to enhance spatial perception. The land-based sensor platform can be seen in Fig.~\ref{fig:campaigns}(b).

\emph{Mission Goals:} (i) Generate a LiDAR-enhanced dataset to support mapping for autonomous docking of harbour busses; (ii) Create a multi-viewpoints dataset to instruct research on cyber-resilient remote navigation of commercial vessels combining the shipborne and land-based sensor platforms.

\emph{Results:} The dual‑platform setup increased logistical complexity and required coordinated land–vessel operation, with deployment taking approximately one hour.
To prevent bandwidth conflicts, LiDAR data were collected on a separate network, and ship and shore recordings were synchronized post‑hoc using GNSS timestamps.
The resulting dataset supported LiDAR‑based quay mapping and detection~\citep{aggerholmSLAMAutonomousDocking}, as well as camera‑based GNSS integrity monitoring for remote navigation~\citep{Louedec_2025}.

\subsection{Mapping of maritime critical infrastructure}
A reduced-version sensor platform was developed for deployment on an Unmanned Surface Vehicle (USV), packaged in a splash‑proof payload module and deployed during multiple campaigns in Danish harbours and marinas. The sensor platform can be seen in Fig.~\ref{fig:campaigns}(c).

\emph{Mission Goal:} (i) Collect synchronized acoustic data from one or more SONAR devices below the water surface and LiDAR data above the surface, to enable the creation of unified 3D ``seabed‑to‑sky'' maps, and support research on localization in GNSS contested environments.

\emph{Results:} Synchronized LiDAR, IMU, and SONAR data were recorded, with a wireless land‑based link enabling real‑time monitoring. LiDAR measurements were processed using smoothing‑and‑mapping to map the above-surface domain and provide odometry, while SONAR data were integrated into the map concurrently, resulting in a consistent seabed-to-sky map ~\citep{thompson2024autonomous, lundMultiDomainSLAMGNSS}.

\section{Discussion and conclusions}
Advancing autonomous ship navigation depends on understanding vessel–environment interactions. A modular framework for sensor integration and data collection is then paramount to advance current capabilities by enabling rapid deployment, and multimodal data acquisition across multiple vessels.

From a data-centric perspective, the presented blueprint facilitates the creation of structured, multimodal datasets that are vital for perception, situational awareness, and decision-making in autonomous systems. The synchronization of diverse sensor modalities supports the generation of rich, annotated datasets aligned with the principles of open and reproducible research.
From a control-systems perspective, the platform is an enabling infrastructure: recorded datasets support replay-based evaluation of state-estimation and perception pipelines in addition to hardware-in-the-loop validation of autonomy components. This reduces the dependency on repeated sea trials and enables systematic benchmarking.

Through the development and extensive deployment of our proposed sensor platform blueprint across varied real-world scenarios, we have identified and addressed practical challenges related to sensor placement, network stability, and environmental changes. Such factors are often overlooked in short-term campaigns or simulations. These insights are critical for both short- and long-duration data collection efforts, where consistent and reliable data streams are key for algorithm development.
The presented campaigns, which spans multi-week ferry crossings, repeated short-duration harbour deployments, and USV operations across varied environments, provide direct operational evidence of this reliability, demonstrating sustained performance across vessel types.

Looking ahead, this blueprint provides a robust foundation for integrating higher-level autonomy modules, including COLREGs-compliant navigation, semantic scene understanding, and AI-based decision support. The adaptability of the platform, demonstrated across different vessels and sensor configurations, underscores its potential for broad deployment and scalable data collection. 

Future work will focus on systematically comparing different configurations on the same vessel to identify how the change in instrumentation can improve situational awareness and algorithmic performance. Synchronized, multimodal datasets captured in a structured format enable researchers to recreate operational scenarios, evaluate perception and sensor‑fusion algorithms, and refine decision‑making without relying solely on repeated real-world trials. This is particularly valuable for training machine learning models and simulating edge cases - such as adverse weather conditions or illumination changes - that are difficult to reproduce in controlled environments.

Ultimately, the proposed sensor platform blueprint represents an important step toward robust, scalable maritime autonomy research infrastructure, supporting practical deployments and the development of high-quality, structured datasets essential for advancing perception and decision-making algorithms.

\begin{ack}
The authors are grateful to Molslinjen for allowing the data collection campaign onboard the ferry \emph{Aurora}. Additionally, the authors would like to thank Movia for access to the Copenhagen harbour buses for data collection.
\end{ack}

\section*{DECLARATION OF GENERATIVE AI AND AI-ASSISTED TECHNOLOGIES IN THE WRITING PROCESS}

During the preparation of this work, the authors used M365 Copilot to improve phrasing of individual sentences and paragraphs. All content was subsequently reviewed and edited by the authors, who take full responsibility.

\bibliography{ifacconf}

@inproceedings{blanke2024greenhopper,
	title        = {GreenHopper: The Danish spearhead towards autonomous waterborne mobility},
	author       = {Blanke, Mogens and Hansen, P Nicholas and Dittmann, Kjeld and Enevoldsen, Thomas T and Dagdilelis, Dimitrios and Sch{\"o}ller, Frederik ETS and Plenge-Feidenhans, Martin K and Becktor, Jonathan and Papageorgiou, Dimitrios and Galeazzi, Roberto},
	year         = 2024,
	booktitle    = {Journal of Physics: Conference Series},
	volume       = 2867,
	pages        = {012035},
    doi          = {10.1088/1742-6596/2867/1/012035}
}

@inproceedings{schmidt2025gabor,
	title        = {Towards better data variation in maritime perception datasets},
	author       = {Schmidt, Jonathan Eichild and Rokseth, Børge and Schliemann-Haug, Asger C. and  Galeazzi, Roberto},
	year         = 2025,
	booktitle    = {Proceedings of 16th IFAC Conference on Control Applications in Marine Systems, Robotics and Vehicles},
    doi = {10.1016/j.ifacol.2025.11.738}
}

@inproceedings{mirza_robustness_2021,
	title        = {Robustness of Object Detectors in Degrading Weather Conditions},
	author       = {Mirza, Muhammad Jehanzeb and Buerkle, Cornelius and Jarquin, Julio and Opitz, Michael and Oboril, Fabian and Scholl, Kay-Ulrich and Bischof, Horst},
	year         = 2021,
	booktitle    = {2021 {IEEE} International Intelligent Transportation Systems Conference ({ITSC})},
	pages        = {2719--2724},
	doi          = {10.1109/ITSC48978.2021.9564505}
	}

@misc{dagdilelis2022,
	title        = {Cyber-resilience for marine navigation by information fusion and change detection},
	author       = {Dagdilelis, Dimitrios and Blanke, Mogens and Andersen, Rasmus H. and Galeazzi, Roberto},
	year         = 2022,
	doi          = {10.1016/j.oceaneng.2022.112605}
}

@misc{eide2025,
	title        = {The Autonomous Urban Passenger Ferry milliAmpere2: Design and Testing},
	author       = {Eide, Egil and Breivik, Morten and Brekke, Edmund F. and Eriksen, Bjoern-Olav H. and Wilthil, Erik and Helgesen, Oeystein K. and Thyri, Emil H. and Veitch, Erik and Alsos, Ole Andreas and Johansen, Tor Arne},
	year         = 2025,
	doi          = {10.1115/1.4067370}
}

@misc{thombre2022,
	title        = {Sensors and AI Techniques for Situational Awareness in Autonomous Ships: A Review},
	author       = {Thombre, Sarang and Zhao, Zheng and Ramm-Schmidt, Henrik and Vallet García, José M. and Malkamäki, Tuomo and Nikolskiy, Sergey and Hammarberg, Toni and Nuortie, Hiski and H. Bhuiyan, M. Zahidul and Särkkä, Simo and Lehtola, Ville V.},
	year         = 2022,
	doi          = {10.1109/TITS.2020.3023957}
}

@inproceedings{grayver2024,
	title        = {Position and Navigation Using Starlink},
	author       = {Grayver, E. and Nelson, R. and McDonald, E. and Sorensen, E. and Romano, S.},
	year         = 2024,
	booktitle    = {2024 IEEE Aerospace Conference},
	volume       = {},
	number       = {},
	pages        = {1--12},
	doi          = {10.1109/AERO58975.2024.10521263},
}

@article{Schmidt_2025online,
	title        = {Online performance evaluation of vision-based perception system for autonomous ship navigation},
	author       = {Schmidt, Jonathan Eichild and Rokseth, Børge and Galeazzi, Roberto},
	year         = 2025,
	month        = {oct},
	journal      = {Journal of Physics: Conference Series},
	volume       = 3123,
	number       = 1,
	pages        = {012025},
	doi          = {10.1088/1742-6596/3123/1/012025}
}

@article{Louedec_2025,
	title        = {Increasing the resilience of MASS remote pilot’s situational awareness using land-based sensors},
	author       = {Louédec, Morgan and Boudot, Solène and Papageorgiou, Dimitrios and Galeazzi, Roberto},
	year         = 2025,
	journal      = {Journal of Physics: Conference Series},
	volume       = 3123,
	number       = 1,
	pages        = {012065},
	doi          = {10.1088/1742-6596/3123/1/012065}
}

@article{chung2023pohang,
	title        = {Pohang canal dataset: A multimodal maritime dataset for autonomous navigation in restricted waters},
	author       = {Chung, Dongha and Kim, Jonghwi and Lee, Changyu and Kim, Jinwhan},
	year         = 2023,
	journal      = {The International Journal of Robotics Research},
	volume       = 42,
	number       = 12,
	pages        = {1104--1114},
    doi = {10.1177/02783649231191145}
}

@inproceedings{choi2025polaris,
	title        = {Polaris dataset: A maritime object detection and tracking dataset in pohang canal},
	author       = {Choi, Jiwon and Cho, Dongjin and Lee, Gihyeon and Kim, Hogyun and Yang, Geonmo and Kim, Joowan and Cho, Younggun},
	year         = 2025,
	booktitle    = {2025 IEEE International Conference on Robotics and Automation (ICRA)},
	pages        = {13626--13632},
    doi={10.1109/ICRA55743.2025.11128583}
}

@article{jang2025moana,
	title        = {MOANA: Multi-radar dataset for maritime odometry and autonomous navigation application},
	author       = {Jang, Hyesu and Yang, Wooseong and Kim, Hanguen and Lee, Dongje and Kim, Yongjin and Park, Jinbum and Jeon, Minsoo and Koh, Jaeseong and Kang, Yejin and Jung, Minwoo and others},
	year         = 2025,
	journal      = {The International Journal of Robotics Research},
    doi          = {10.1177/02783649251354897},
}

@article{Macenski_2022,
	title        = {Robot Operating System 2: Design, architecture, and uses in the wild},
	author       = {Macenski, S and Foote, T and Gerkey, B and Lalancette, C and Woodall, W},
	year         = 2022,
	journal      = {Science Robotics},
	volume       = 7,
	number       = 66,
	doi          = {10.1126/scirobotics.abm6074},
	issn         = {2470-9476}
}

@inproceedings{defilippo-robowhaler,
	title        = {Robowhaler: A robotic vessel for marine autonomy and dataset collection},
	author       = {DeFilippo, Michael and Sacarny, Michael and Robinette, Paul},
	year         = 2021,
	booktitle    = {OCEANS 2021: San Diego--Porto},
	pages        = {1--7},
	doi          = {10.23919/OCEANS44145.2021.9705871}
}

@inproceedings{wirbel2025data,
	title        = {Unlock AI Vessel Navigation Assistants: A Community Call for Open Data},
	author       = {Wirbel, Julius and Clemmensen, Line and Galeazzi, Roberto},
	year         = 2025,
	booktitle    = {Proceedings of 16th IFAC Conference on Control Applications in Marine Systems, Robotics and Vehicles},
    doi = {10.1016/j.ifacol.2025.11.743},
}

@inproceedings{labust_sensorplatform,
	title        = {Design and Development of a Boat-Mountable Sensor Rack for Maritime Perception and Data Acquisition},
	author       = {Obradović, Juraj and Fabijanić, Matej and Lovrić, Josip and Kapetanović, Nadir and Na\dj, {\DJ}ula and Ferreira, Fausto  and Nikola Mišković},
	year         = 2025,
	booktitle    = {Proceedings of 16th IFAC Conference on Control Applications in Marine Systems, Robotics and Vehicles},
    doi = {10.1016/j.ifacol.2025.11.640},
}

@inproceedings{thompson2024autonomous,
	title        = {Autonomous Inspection and Data Fusion for Maritime Critical Infrastructures},
	author       = {Thompson, Fletcher and Hansen, Peter Nicholas and Galeazzi, Roberto and Palma, Marco and Brock, Andreas Libonati and Mariani, Patrizio},
	year         = 2024,
	booktitle    = {2024 27th International Conference on Information Fusion (FUSION)},
	pages        = {1--8},
    doi={10.23919/FUSION59988.2024.10706512}}

@article{lundMultiDomainSLAMGNSS,
	title        = {Towards Multi-Domain SLAM in GNSS Denied, Maritime Urban Environments},
	author       = {Lund, Aimas and Hansen, Peter Nicholas and Thompson, Fletcher and Prabowo, Yaqub A and Galeazzi, Roberto},
	year         = 2025,
	journal      = {Proceedings of 16th IFAC Conference on Control Applications in Marine Systems, Robotics and Vehicles},
    doi = {10.1016/j.ifacol.2025.11.622}
}

@article{aggerholmSLAMAutonomousDocking,
	title        = {{{SLAM}} for {{Autonomous Docking}}: {{A Case Study}} of {{Copenhagen}}’s {{Harbour Buses}}},
	author       = {Aggerholm, Oda B and Prabowo, Yaqub A and Hansen, Peter N and Galeazzi, Roberto},
	year         = 2025,
	journal      = {Proceedings of 16th IFAC Conference on Control Applications in Marine Systems, Robotics and Vehicles},
    doi = {10.1016/j.ifacol.2025.11.626},
}

\end{document}